# Artificial Intelligence and Economic Theories

Tshilidzi Marwala and Evan Hurwitz

University of Johannesburg

**Abstract**

The advent of artificial intelligence has changed many disciplines such as engineering, social science and economics. Artificial intelligence is a computational technique which is inspired by natural intelligence such as the swarming of birds, the working of the brain and the pathfinding of the ants. These techniques have impact on economic theories. This book studies the impact of artificial intelligence on economic theories, a subject that has not been extensively studied. The theories that are considered are: demand and supply, asymmetrical information, pricing, rational choice, rational expectation, game theory, efficient market hypotheses, mechanism design, prospect, bounded rationality, portfolio theory, rational counterfactual and causality. The benefit of this book is that it evaluates existing theories of economics and update them based on the developments in artificial intelligence field.

## 1.1 Introduction

"Workers of the world unite, you have nothing to lose but chains" so said Karl Marx (Marx, 1849). Despite what many Marxists claim, he never foretold the advent of artificial intelligence, otherwise he would probably have said "Artificial intelligent machines of the world unite, you have nothing to lose but chains". But what Marx realized was that the principal agent of work is man. Man is the invisible hand that drives the economy as observed by Adam Smith (Smith, 2015). The economy was by man and about man but the theories that explained the economy did not quite match the behaviour of a man. For this reason the rational man is indeed irrational and his irrationality permeates every aspect of life including the very concept we call the economy.

Homo-Sapiens have been around for hundred thousand years and throughout their existence and even from their forbearers have inherited certain traits and behaviours that influence them even today (Harari, 2014). Some of these traits and behaviours include greed, fear, bias and social structure. All these traits are still with us today because of one and only one reason and that it because they all have given us an evolutionary advantage. Of course, this might change in the future depending on the change of environment and therefore these traits might depart from human beings. All these traits influence our decision making and the rationality thereof. Herbert Simon calls the idea of making decisions making with all these constraints e.g. processing power of our brains, incomplete information and human behaviour, bounded rationality. Our rationality is bound to all these constraints but what will happen when machine replaces humans. Is the rationality of machines bound? (Simon, 1991) Are the bounds of rationality bigger in humans than machines? These are some of the questions that this book seeks to answer.

Machines are now part of everyday decision making. They are becoming more intelligent due to a technology called artificial intelligence (AI). Alan Turing surmised that machines are intelligent if and only if when we interact with them we cannot tell whether we are interacting with a man or a machine (Traiger, 2000). This is what is called a Turing test. No machine has passed this Turing test over an extended levels of man-machine interaction. But this does not limit artificial intelligence and make machines incapable of solving complex problems.

This paper describes man-machine interaction and its impact on some of the big ideas in economics. Every economic epoch had its own theories or thinking. Some of these theories stood the test of time but some have fallen by the wayside. The biggest event in history of economics is the history of industrialisation and all its various stages. The first industrial revolution happened in 1874 in England. What is not clear is why it did not happen in Asia especially India or China as probabilistically these two regions had higher population densities. What was the catalyst that caused the first industrial revolution? In the 17 century lived a man in England called Isaac Newton who was educated at Trinity College Cambridge (Newton, 1726). Legend claims that unlike many people who had lived before him and had witnessed an apple falling, he asked: "Why did the apple fall?" And from this he discovered gravity an intriguing concept which was only given an explanation by a German/Swiss/American scientist Albert Einstein some hundreds of years later. Newton, furthermore, came with what is now called the laws of motion which stated that objects will continue at rest or keep on moving until they are moved or stopped respectively. Furthermore, he observed the relationship between force, mass of an object and its acceleration. This thinking of Newton, made us understand movement and became the catalyst or DNA for the first industrial revolution. This gave us steam engines, trains and mechanical machines for production. From this era economic principles such as Marxism, Adam Smith's invisible hand as well as David Ricardo's labor theory of value and the principle of comparative advantage were conceived (de Vivo, 1987).

Then in the 19th century came a British man called Michael Faraday who performed crucial experiments which were later interpreted by James Clerk Maxwell through his beautiful mathematical equations (Maxwell, 1873; Agassi, 1971). Michael Faraday observed that if you have a magnet and you put next to it a wire that conducts electricity and you move the wire, then electricity flows in that conductor. This is wizardry beyond miracles of biblical proportions. Even today we generate electricity, perhaps with the exception of solar energy and few others, using this technique. Conversely, Faraday observed that again with a magnet and a conducting wire and you force electricity through the wire, then the wire moves and this was the birth of the electric motor that still moves our assembly lines. These events were the DNA for the second industrial revolution. From this era economic principles such as mass production and Keynesian economics were introduced.

Then in the second half of the 20th century John Bardeen, Walter Brattain, and William Shockley discovered the transistor (Amos and James, 1987). It is based on semiconductors which are objects that conduct electricity under certain conditions. The transistor is the catalyst for the electronic age that gave us computers, cell phones, information technology and automation in our factories. It is the DNA of the third industrial revolution. It was in the era that powerful economic concepts such as market efficiency where introduced prospect theory.

The era we are living is the fourth industrial revolution. This is an era of intelligent machine (Marwala, 2007; 2009; 2010; 2012; 2013; 2014; 2015; Marwala and Lagazio, 2012; Marwala et al, 2016). The DNA of the fourth industrial revolution is AI. It will touch all aspects of our lives. We will have robotic cops to protect us, robotic doctors to assist us with medical issues, all our vital organs will be monitored real time to extend human lives, driverless cars and aircraft as well as human empty factories as human will be replaced. What will happen to economics? Already we know that the stock market no longer has crowds of people shouting a price of stocks because artificial intelligent software are doing the work. This paper explores how economic theories that have guided decision makers in the past will have to change or adapted in the light of artificial intelligent capabilities.

**1.2 Economics and Economic Theory**

Economics began when man started battering for exchanging goods. This was mainly based on the reality that man could not produce all he wanted. For example, suppose we have a man called Peter who produces maize and another called John who produces peanuts. Then Peter will give half of his maize for half of John's peanuts. If we include a third person Aiden who produces wheat, then Peter takes a third of his maize and gives it to John in exchange of a third of his peanuts, gives another third to Aiden in exchange of his third of wheat. It is evident that this becomes a complicated problem quickly as more and more people exchange goods.

To facilitate this process an additional entity called money comes into the picture for one and one reason only and that is to simplify the exchange of goods and services. Well not quite because money on its own is worthless. It becomes useful when it has a property called trust. Trust is what makes money useful. You take away the trust as Robert Mugabe did in Zimbabwe you collapse money as it happened in when Zimbabwean dollar was replaced by an American dollar as a legal currency. In this example there are factors of production that makes this transaction happen and these are labour to plough, capital and here we are talking about seeds and land to plough these seeds on. The study of all these activities is called economics. The generalization of this activities such as the philosophy on how to price these goods is called economic theory. Generalization is of course a big concept. For example if most of the time if it is cloudy and humid it rains then one can generalize and say there is a natural law that states that whenever it is cloudy and humid then it will probably rain.

If money is now the object to be used to exchange goods how do we put a price to peanuts, maize and wheat? Well it depends on your ideology. In the strictly planned economy the authority determines the price whereas in the market economy it is determined by the demand and supply characteristics whereas in a market in Delhi you negotiate a price and it is not fixed. One thing for sure, the producers of goods and services are only willing to sell their goods at a price which is higher than the cost of producing these goods. This paper will explore how artificial intelligence change pricing and the theory of pricing.

On the theory of pricing one looks at the supply and demand and in our case we will use peanuts. When the supply and demand are equal this is called equilibrium. Equilibrium is a balance of all economic forces and in our case these forces are the supply and demand of goods and services. At the point of equilibrium there is no incentive to change because changing does not advantage any agent. With the advent of artificial intelligence, we can now employ multi-agent system to simulate an economic process and observe the point of equilibrium thereby assisting in matters such as pricing.

The production of peanuts, wheat and maize can be industrialized using tractors. In this situation the owners of capital do not necessarily have to work, the workers have to work. In the worst case scenario, these workers can be so exploited that they are paid a bare minimum wage. From a situation like this in the major industrial towns of Britain, Karl Marx looked at this and crafted an ideology where the workers become the owners of capital through a revolution. He went further and stated that "To each according to their needs and from each according to their abilities". This of course was against human nature and could not be enforced without the violence of Lenin and Stalin.

Adam Smith looked at organization of the economy and observed that as individuals pursue goals of maximizing their returns the greater good of increasing production and income are achieved. He called the driver of this greater good the invisible hand. The concept of maximizing the greater good is what is called utilitarianism and was proposed by John Stuart Mill (Mill, 2001). When these owners of capital and workers pursuing their individual wants the emergence outcome if the greater good for society. Of course with the advent of multinationals that are big, and the influence of finance and advertising this is not necessarily the case. In fact as Thomas Picketty observes in his seminal book *Capital in the 21$^{st}$ Century* that inequality is still a major problem (Piketty, 2014). As Nobel Laureate Joseph Stiglitz in his book *The Price of Inequality* observes that inequality stifles economic growth (Stiglitz, 2012). The idea of the invisible hand is phenomenon that is observed even in biological systems. For example, the swarming of pigeons which is illustrated in

Figure 1 is an example of the invisible hand because there is no leader. Each bird looks at what its neighbours as well as the group are doing and follows and what emerges is a coordinated movement of the swarm towards the food source. This phenomenon has been translated into an artificial intelligence algorithm called particle swarm optimization.

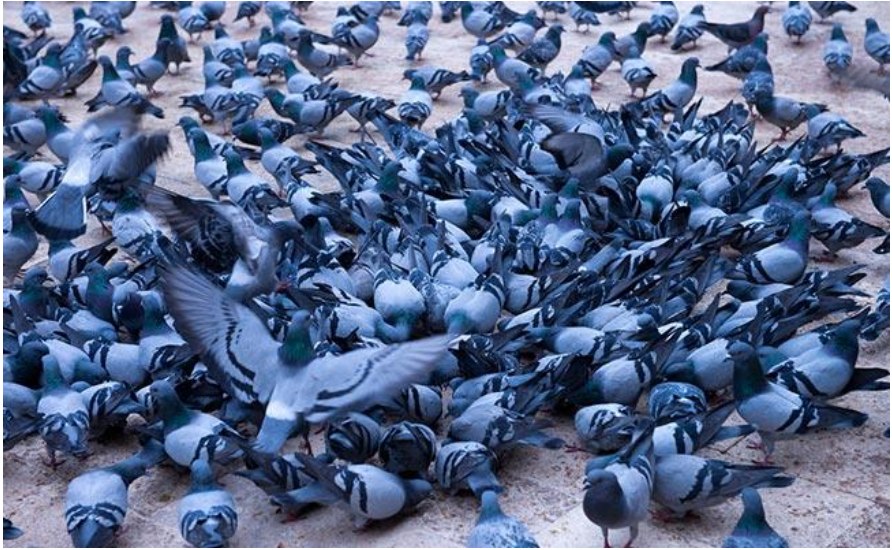

Figure 1 The swarming of pigeons

The other important factor is the concept of value. In our example, between peanuts, wheat and maize which one is more valuable? Should we break them into their chemical components and price each compound? The answer is very similar to what William Shakespeare said about beauty: "It is on the eyes of the beholder". It depends on an individual's frame of reference and the reason why these three farmers will exchange goods is because of the differences in the perception of values. Even within a single product there is a difference in the perception of value for example for Peter only 1/3 of maize is more valuable than wheat and peanuts and because of this fact Peter is only willing to trade the 2/3 of his maize. The concept of asymmetry of value and its principal importance in trade will be discussed in detail later in this paper and it will be established that AI reduced information asymmetry.

In an example we have given, Peter may go and create a company Peter Incorporated. The company will have limited liability and the theory behind this is that in order to encourage people to form companies a legal framework is established to limit personal liability in situation when the company runs into difficulties. The company is therefore established as a legal entity independent of its owners and therefore has rights and obligations. Peter may then decide to list the company in the local stock exchange to raise funds for the company and by so doing distributes shareholding to members of the public.

There is whole discipline on how to predict the future price of stocks and there are two general ways of doing this and these are through fundamental and technical analyses. Technical analysis is a mathematical exercise where one looks at the data of the performance of the stock and make decisions on whether to buy such a stock or not. The problem with this is that people buy stock for no reason except that it is going up irrespective of how solid is the company with that particular share price. The dotcom bubble that busted in the early 2000 was largely based on this where a company with a shaky business model was listed in the stock market and its share price skyrocketed and then crashed to the ground (Hamilton, 1922; Siegel, 2008). The other way of predicting the share price is by looking at the underlying business activities of companies and then decide whether to buy the share price or not. Both these approaches have merit and today with the advent of big data and AI technical analysis is augmented with data from the internet whether twitter or searching through the internet to find out what the market sentiments, including quantification of the fundamentals, on the particular stock are. These will be discussed in this paper in a section on quantitative analysis.

Another theory that has emerged from the stock market is the theory of the efficient market hypothesis. This theory basically states that it is impossible to beat the market because the share price reflects all elements that are relevant. This theory states that even though there might be instances where a share price might not already incorporate all relevant information it then self corrects to reach an equilibrium point when the markets are efficient. Much of the information in the stock are based on human decisions which are unpredictable and in most cases irrational. For a share price to incorporate all these decisions and their future intentions, is almost importable. Irrational people make decisions in the market these result with irrational markets. But now we know that much of the decisions in the market are made by artificially intelligent machines and the volume of these will expand into the future. Therefore, as these happen then the markets become more efficient and this is discussed in detail in this paper.

Peter, John and Aiden when they make their decisions on when to plan, how to plant, what fertilizers to use, how to trade, at what price and who to seek advice from, they require certain attributes to make these decisions. The first attribute is information and this is often limited, not wholly accurate and often missing. The second attribute is sufficient mental capability to make sense and process the information which is often less than perfect. This is the concept that Economics Nobel Prize Laurate and AI expert Herbert Simon called bounded rationality. He further observed that in such a situation, Peter, John and Aiden, will have to satisfice meaning obtaining satisfactory and sufficient solution. Now if we add AI into this decision making process what happens to missing data and processing of information?

All the players in the market have to make decisions, whether how much peanuts, maize and wheat to be planted. They make these decisions individually and/or collectively. The question that has been asked for many generations is: How do people make their choices? The theoretical answer is that they make their choices based on their desires to maximize their utilities. There is a difference between a desire to maximize utility and actually been able to make a rational decision. Studies have shown that the theory of rational choice is not what drives decision making. In fact Kahneman in his book *Thinking fast and slow* demonstrated that people make decisions based on their aversion to loss (Kahneman, 2011; Anand, 1993).

Now with the advent of artificial intelligence, is it not possible to design machines that are able to make decision based on rational choice theory? Doesn't artificial intelligence in decision making implies that decisions are more and more going to be based on the theory of rational choice rather than aversion to loss? Another closely aligned matter is the issue of rational expectation theory which states that people make decisions based on future outlook, information available and past experience. AI machines look at available information and past information to make a decision and excludes individual perception of the future. Therefore, the degree of uncertainty in AI made decisions is lower than that made by human beings. For example, if one asks human beings to make decisions at different times they will change their minds depending on all sorts of irrelevant factors.

As people make their decisions on what to do, there is a framework that has been developed to aid with that and this is called game theory. Game theory is a mathematical framework which assists us in making optimal decisions given the fact that we do not know what other people's choices will be but their choices influence the outcome. There are many types of games e.g. zero sum games where the gain for one player is the loss by another player. In game theory there are a number of assumptions that are made and these include the fact that players are rational, and there are sets of rules and the games are played till Nash equilibrium is reached (Nash, 1950). Nash equilibrium is a position where each player cannot improve his utility by playing further. With the advent of artificial intelligence and in particular multi-agent systems the whole theory of games can be better implemented and for much complicated games for the benefit of better decision making.

Another framework that has been developed is the framework of mechanism design which in our example can be used to design a market for the peanuts industry. Mechanism design won Leonid Hurwicz, Eric Maskin, and Roger Myerson a Nobel Prize in Economics (Hurwicz et al, 1975)). Mechanism design is in essence reverse game theory where instead of players, rules and then identifying equilibrium, here you have the desired equilibrium and players and you want to design the rules so that the desired equilibrium can be reached. This is in effect a control problem where the system is designed to achieve a certain desired outcome. For an example, when an aircraft is flying over a highly turbulent environment then the autopilot system identifies the appropriate speed, angle of attach and altitude to achieve maximum comfort for the passengers.

Suppose a trader has $1 million is faced with making a choice on buying stocks from a list of 100 stocks. The trader, has two options either use $1 million to buy one stock or can buy a basket of stocks. The process of choosing the optimal basket of stocks is called portfolio optimization. The optimal basket today is not the optimal basket tomorrow and how then does he optimize this? Perhaps he can use a time window, say 2 months, to find the average optimal basket. All these variables, e.g. time window, selection that forms a basket, how large should the basket be, are all unknown variables. Nobel Prize Laureate, Harry Markowitz, proposed what is called portfolio optimization theory to select this basket of stocks. However, his theory is based on a number of assumptions including the fact that the character of the basket of stocks does not change, something that is called stationarity in mathematical nomenclature (Markowitz et al., 1952). With the advent of AI, one is now able to apply this technology to the theory of portfolio optimization. For example, how do we solve portfolio optimization problem using the theory of ant colony intelligence which is based on depositing pheromones as they move and following the path with the strongest pheromones, and then converging on an equilibrium path which happens to be the shortest distance between the food source and ants nest?

The theories of factual and counterfactuals are at the heart of economic theory. For example, if the farmer Aiden plants wheat and instead planted maize, will he have obtained a better return on investments? These two scenarios, planting wheat, factual (planting wheat because it happened) compared to the counterfactual (planting maize because it did not happen). The economists have a beautiful word for this and they call this opportunity costs (Henderson, 2008). The question that needs to be answered in this thesis is whether we can use AI, with its predictive, adaptive and learning capabilities, to estimate the counterfactuals and thus economic costs. If we are able to quantify the opportunity cost it means this can be integrated into decision making to make choices that maximize return on investments.

Another issue that is closely related to the issues of factual and counterfactual is the concept of causality (Gujarati and Porter, 2009). Nobel Laureate Clive Granger thought he had pinned down this matter when he proposed Granger causality which even though is useful particularly in economics is not causality at all but some measure of correlation (Dawn, 2009). We know that detecting causality on static data is difficult and that only experimentation is the most reliable way of detecting causality. A causes B if and only if there is a flow of information from A to B. How do we then detect such flow of information? Alternatively, we can frame causality in terms of factual and counterfactual by posing the following problem instead: A causes B (factual) if when A doesn't happen then B does not happen. This paper illustrates the centrality of causality in understanding economic problems and proposes several ways in which to understand economics. We explore the theories of rational choice and rational expectations within the dimensions of causality.

## 1.3 Artificial Intelligence

In order to rewrite economic theory in the light of artificial intelligence it is important to understand what artificial intelligence is. AI is made out of two words, artificial and intelligence and thus it is intelligence that is artificially made. Intelligence is the ability to make sense of information beyond the obvious (Marwala et al, 2006; Marwala, 2007). There are two types of intelligence in nature and these are individual intelligence and group intelligence. Individual intelligence is intelligence located in a single agent for example the farmer Aiden is individually intelligent, whereas group intelligence is when it is located in a group of agents such as the swarming of pigeons shown in Figure 1 or the school of fish.

Within the field of AI intelligence manifests itself in two ways, specialised intelligence and generalised intelligence. For example, a robot designed at the Korean Advanced Institute of Science and Technology, shown in Figure 2, which drives a car, opens doors, walks and drills holes is demonstrating generalized intelligence. This is because it is able to do multiple independent tasks. A voice recognition software which is able to hear and interpret a person is a specialised robot because it is trained and is able to do one task only.

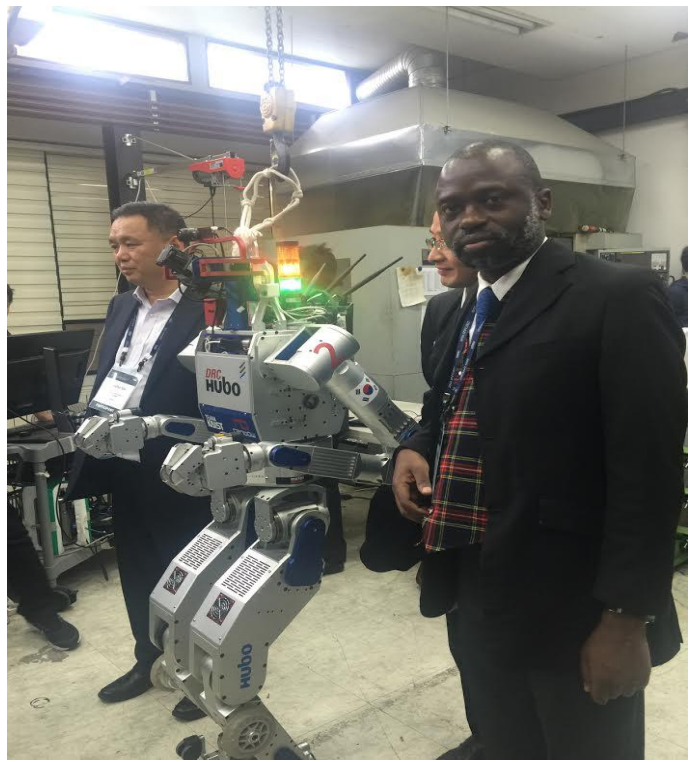

Figure 2. The KAIST humanoid robot amongst human

In artificial intelligence, there are various capabilities that have been invented and these are how to make machine learn, optimize, predict, adapt and interact. On learning, the source of knowledge is how a human brain functions and from this a neural network was invented which is able to take information from the world and interpret it. For example, a neural network is able to take an x-ray picture of the individual's lung and then establish whether the individual has pulmonary embolism or not. Neural network is also able to take an EEG signal from an individual and establishes whether the individual will be experiencing epileptic activities or not in the near future.

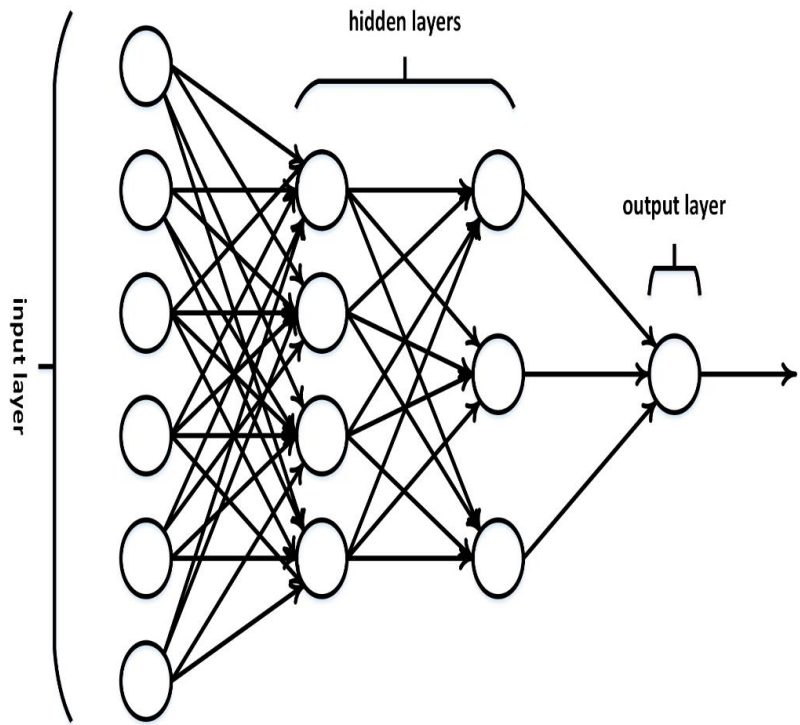

Figure 3 An example of a neural network

In Figure 3, we have input and output layers. The input layer has the variables that are to be used as a basis for prediction whereas the output layer is what is to be predicted. An example, is a study on using the income of parents to predict the educational outcome of the children as measured by whether they go to college or not. Here the income of the parents will be in the input layer whereas the educational outcome of the children will be in the output layer. The hidden layers constitute the mechanism or the engine which facilitates the translation of input data into the output. This engine consists of neurons (inspired by natural neurons) and the identity of these neurons which are represented by numbers is obtained from the observed historical data. The identification of these neuron is obtained through the process called optimization. The interesting aspect of this setup is that the prediction is essentially derived from the historical observation and thus confirms the quote that: "Predicting the future is nothing more than the rearrangement of information that currently exists"

This are tasks that normally done by trained doctors. In fact, doctors here use what is called in economic theory of rational expectation by taking current and previous histories and their predictions of the future to make decisions. What AI does is not necessarily to change the predicted diagnosis but to bring more consistency which in statistically terms called reduction of the degree of uncertainty. There are other AI techniques that have been designed for learning and these are fuzzy logic which brings precision to the spoken words and use these to make decisions through a computer and support vector machines which are based on increasing the dimension of the space to better differentiate attributes in objects.

The second aspect of AI is optimization. Optimization is a process of identifying the lowest or a highest point in a problem. For example, the process of identifying who is the shortest man in the world is an optimization problem. Using this example, the optimization algorithm is a computer or mathematically based procedure of identifying who the shortest person in the world is. Another example if you are designing an aircraft you might intend to make all your choices e.g. material to use and shape of the plane to result in the lowest amount of aeroplane weight. The parameters that you will play around are called design variables. Nature is full of optimization processes and one example is the process of natural evolution. This controversial theory, especially amongst the religious right, is a process in which species adapt to their environment and they achieve this through four drivers and these are mutation, random alteration of genes, crossover, mixing of genes, as well as reproduction of those individuals that are most adaptive to their environment. This natural optimization process has been codified to produce a computer based process called genetic algorithm.

The other aspect of AI is adaptation which is an optimization problem. For example, if one was to create a predictive system that predicts whether it is going to rain or not. The first level of optimization is to set up parameters that will make a predictive system predicts whether it will rain or not. The second optimization level ensures that such a predictive system adapts to the evolving environment due to global warming. There have many techniques that have been developed to make adaptation possible and these include fish school algorithm which is inspired by the school of fish behaviour and ant colony optimization which is inspired by the colony of ants.

**1.4 Economic Theories**

*Supply and Demand*

The law of demand and supply is the fundamental law of economic activities. It consists of the demand characteristics of the customer which describes the relationship between price and quantity of goods. For example, if the price of a good is low the customer will buy more goods and services than if the price is high. The relationship between price and the willingness of the customers to buy goods and services is called the demand curve. The other aspect of the demand and supply law is the supply curve which relates the relationship between the price and the quantity of goods suppliers are willing to produce. For example, the higher the price the more the goods and services the suppliers are willing to produce. Conversely, the lower the price the lesser the goods and services the suppliers are willing to produce. The point at which the suppliers are willing to supply a specified quantity of goods and services which are the same as those that the customers are willing to buy is called equilibrium. Artificial intelligence allows companies, such as Amazon, to gather information on the customers such that they are able to create an individual demand and supply curve and therefore price individually based on historical trends. In this regard individuals are not subjected to prices derived from aggregate demand and supply(Marwala and Hurwitz, 2017).

*Rational choice*

Rationality has played a pivotal role in economic theory. Some of the theories that have been used assume perfect rationality. This paper studies the theories of rational choice, preference and utility maximisation. Rational choice theory prescribes that when agents make decisions they do that based on the desire to maximize utility. However, in perfect rationality each move these agents make on building towards the ultimate decision is not necessarily optimal, however, the aggregate of all the moves they make can be rational. A type of artificial intelligence called reinforcement learning can be used as a mechanism for modelling rationality. The concept of local and global optimization can be explored within the context of evolutionary design to study the subject of rationality. Problems where acting rationally locally is vital for the overall global rational solution for example making a stupid move in a game in order to deceive your opponent are studied and the condition in which an irrational technique becomes necessary for achieving global rationality can be derived (Marwala and Hurwitz, 2017).

*Rational expectations*

The theory of rational expectations prescribes that people predict the future based on past and available information and their predictions are not systematically wrong. In this way, predicting the future is nothing more than the arrangements of the information that already exists. The theory of adaptive expectations predict the future based solely on the previous data and this has been found to offer biased conclusions. With the advent of artificial intelligence, data fusion and big data the information that can be used in models to predict the future include mined text data and pictures and this render the predictions of rational expectations more accurate and thus lowers the degree of uncertainties in decision making (Marwala and Hurwitz, 2017).

*Bounded rationality*

The theory of flexibly-bounded rationality is an extension to the theory of bounded rationality. In particular, it is noted that a decision making process involves three components and these are, information gathering and analysis, the correlation machine and the causal machine. Rational decision making involves using information which is almost always imperfect and incomplete together with some intelligent machine which if it is a human being is inconsistent to make decisions. In the theory of bounded rationality, this decision is made irrespective of the fact that the information to be used is incomplete and imperfect and that the human brain is inconsistent and thus this decision that is to be made is taken within the bounds of these limitations. In the theory of flexibly-bounded rationality, signal processing is used to filter noise and outliers in the data and the correlation machine is applied to complete missing information and artificial intelligence is used to make more consistent decisions. This results with the bounds prescribed by the theory of bounded rationality expanding and thus increasing the degree of rationality such decisions are made (Marwala and Hurwitz, 2017).

*Asymmetric information*

When agents come together to make decisions it is often the case that one agent has more information than the other and this distorts the market. Often if one agent intends to manipulate a decision in its favour the agent can signal wrong or right information. Alternatively one agent can screen for information to reduce the impact on asymmetric information. With the advent of big data and artificial intelligence signalling and screening has been made easier. The impact of artificial intelligence on the theory of asymmetric information particularly given the fact that screening is made easier due to artificial intelligence and data mining is that it reduces the degree of information asymmetry in the markets and thereby reduces the resultant distortion of the markets by making them more efficient as well as reduces the volume of trades in the market. Therefore, the more artificial intelligent there is in the market the less is the volume of trades in the market, and the overall efficiency of the market is likely to improve over time as the market becomes more saturated with intelligent trading and analysis agents (Marwala and Hurwitz, 2017).

*Pricing*

Pricing theory is a well-established mechanism that illustrates the constant push-and-pull of buyers versus consumers and the final semi-stable price that is found for a given good. The theory accounts very well for the two-person situation of a single buyer and a single seller. This problem becomes exponentially more difficult as more and more buyers and sellers are introduced into a market model, especially if they have strategies to stay ahead of not only their target, but also their competitors. Multi-agent intelligent systems are used to model complex behaviour in each agent, laying the groundwork for creating complex price-finding models based on the introduction or removal of given agents from a large, complex system. These models can be robust and detailed enough to not only capture the steady-state final pricing, but also the dynamics of the price fluctuations that will be of great interest to anyone wishing to model the price movements of goods within the marketplace. The theory of pricing closely linked to the theory of value because the price customers are willing to pay for goods and services is linked to how they value these goods and services. Contrary to what David Ricardo postulated when he proposed the labour theory of value where the value of goods and services is measured by how much labor went into the manufacturing of those goods and services, value is dynamic and artificial intelligence offers a unique opportunity to dynamically estimate those values. Given the fact that the emergence of artificial intelligence and big data in the market place implies individualized supply and demand curves, this implies that pricing is now individualized instead of being derived from aggregate demand and supply curves (Marwala and Hurwitz, 2017).

*Game theory*

Game theory has been used quite extensively in economic problems. In game theory agents with rules interact to obtain pay-off at some equilibrium point often called Nash equilibrium. The advent of artificial intelligence makes the multi-agents game theory possible. The impact of this development is that the implementation of this on issues such as equilibrium and optimal decisions is far reaching. Furthermore, statistical physics is introduced and multi-agent games can now be used introduced to study games (Marwala and Hurwitz, 2017).

*Mechanism design*

In game theory players have rules and pay-off and they interact until some point of equilibrium is achieved. This way we are able to see how a game with sets of rules and a pay-off reaches equilibrium. Mechanism design is the inverse of that, we know what the end-state should look like and our task is to identify the rules and pay-off function which will ensure that the desired end-state is achieved. This is done by assuming that the agents in this setting acts rationally. Deep learning, big data and mechanism offer exciting avenues for economics (Marwala and Hurwitz, 2017).

*Prospect theory*

Prospect theory offers both a fascinating alternative to modelling agent behaviour in a virtual market place, as well as an opportunity for further expanding prospect theory by utilising multiple agents in a larger simulated marketplace utilising multi-agent modelling. The basic premise of prospect theory is that the probabilities of profit / loss have a disproportionately large effect on the behaviour of human agents in an economic system when compared to the actual expected outcome, a premise that is borne out by the results of research in the field of behavioural economics, and one that contradicts the expectation of rationality that is used as a basis of so much economic modelling and theory, especially in relation the assumption of rationality that forms the bedrock of microeconomic modelling. A larger model consisting of intelligent agents that utilise a prospect-theory-based means of determining risk would therefore offer greater accuracy in modelling the actual marketplace than the usual, truly rational agents in more common use. This study surmise that the applicability of prospect theory is determined by the level of artificial intelligence decision makers that are deployed in a given problem (Marwala and Hurwitz, 2017).

*Efficient market hypothesis*

The efficient market hypothesis (in its varying forms) has allowed for the creation of financial models based on share price movements ever since its inception (Fama, 1970). In this paper the base assumptions drawn from root economic principles are brought into question. Their viability for application to the actual marketplace is dissected, elaborating on their successes as well as their shortcomings. The powerful techniques available within the fields of *machine learning* and *multi-agent modelling* are explored, illustrating their potential to forge a new paradigm in which more accurate models can be created upon a basis that presents a truer reflection of the underlying marketplace. In particular the application of multi-agent modelling to the market-modelling problem provides a powerful tool for simulating heterogeneous agents with differing views and objectives in order to create a more realistic market upon which to base our hypotheses. It is surmised that the use of artificial intelligence in the market makes the market more efficient (Marwala and Hurwitz, 2017).

*Portfolio theory*

The basis of portfolio theory is rooted in statistical models based on Brownian motion. These models are surprisingly naïve in their assumptions and resultant application within the trading community. The application of artificial intelligence and machine learning to portfolio theory and consequently to the applications of portfolio theory, portfolio management in particular, have broad and far-reaching consequences. Artificial intelligence techniques allow us to model price movements with much greater accuracy than the random-walk nature of the original Markowitz model, while the job of optimising a portfolio can be performed with greater optimality and efficiency using evolutionary computation while still staying true to the original goals and conceptions of portfolio theory. A particular method of price movement modelling is shown that models price movements with only simplistic inputs and still produces useful predictive results. A portfolio rebalancing method is also described, illustrating the use of evolutionary computing for the portfolio rebalancing problem in order to achieve the results demanded by investors within the framework of portfolio theory (Marwala and Hurwitz, 2017).

*Rationality in facts and counter-facts*

This section introduces the concept of rational counterfactuals which is an idea of identifying a counterfactual from the factual (whether perceived or real), and knowledge of the laws that govern the relationships between the antecedent and the consequent, that maximizes the attainment of the desired consequent (Byrne, 2005). In counterfactual thinking factual statements like: 'Greece was not financially prudent and consequently its finances are in tatters and with its counterfactual being: 'If Greece was financially prudent and consequently its finances are in good shape'. In order to build rational counterfactuals artificial intelligence techniques can be applied to identify the antecedent that gives the desired consequent which is deemed rational (Marwala and Hurwitz, 2017). The rational counterfactual theory can be applied to various problems in economics.

*Quantitative finance*

Quantitative finance has grown with the advents of computing and this growth has accelerated in the last decade with the growth of artificial intelligence. Techniques on how to price risk and forecast the stock price have been enhanced by the use of artificial intelligence. Subjects such as evolution, deep learning and big data are changing the effectiveness of quantitative finance.

*Economic Causality*

Causality is a powerful concept which is at the heart of markets. Often one wants to establish whether a particular attribute causes another. Often as human beings we have perceived causality through correlation. Because of this fact, causality has often been confused for correlation. The evolution of causality including the influential work of David Hume and its relevance to economics and finance is an important area of study (Marwala and Hurwitz, 2017). Various concepts and models of causality such as transmission, Granger and Pearl models of causality can be improved by the use of artificial intelligence. The transmission model of causality states that for causality to exist there should be a flow of information from the cause to the effect. Bayesian inference as a mechanism of modelling causality can be used to study the link between cigarette smoking and lung cancer.

## 1.5 Conclusion

This paper went through some of the big ideas that have been developed in the study of economics. These include concepts such as Marxism, the theory of invisible hand, rational expectations, rational choice, mechanism design and game theory. Furthermore, this paper discussed how some of these will change as man is replaced by an artificial intelligent machine as a principal agent of economic decision making.